%% file: main.tex
\documentclass{article}

 \usepackage[main, final]{neurips_2026_x}

\usepackage[utf8]{inputenc} 
\usepackage[T1]{fontenc}    
\usepackage{hyperref}       
\usepackage{url}            
\usepackage{booktabs}       
\usepackage{amsfonts}       
\usepackage{nicefrac}       
\usepackage{microtype}      
\usepackage{xcolor}         

\usepackage{graphicx}
\usepackage{gensymb}

\usepackage{xcolor}

\usepackage{subcaption}

\usepackage{kotex}
\usepackage{color}
\usepackage{array}
\usepackage{hyperref}

\usepackage{bm}
\usepackage{nth}
\usepackage{setspace}

\newcolumntype{C}[1]{>{\centering\let\newline\\\arraybackslash\hspace{0pt}}m{#1}}
\newcolumntype{L}[1]{>{\let\newline\\\arraybackslash\hspace{0pt}}m{#1}}

\usepackage{rotating} 

\usepackage{amsmath}

\usepackage{algorithm}
\usepackage{mathtools, nccmath}
\usepackage{booktabs}

\usepackage{graphicx}
\usepackage{textcomp}
\usepackage{booktabs}
\usepackage{multirow}

\usepackage{tabularx}
\usepackage{makecell}
 
\def\BibTeX{{\rm B\kern-.05em{\sc i\kern-.025em b}\kern-.08em
    T\kern-.1667em\lower.7ex\hbox{E}\kern-.125emX}}

\title{Do Vision Models Encode Object-Level Semantic Relatedness? A Cognitive Psychology-Inspired Benchmark}

\author{%
  Hansang Lee \\
  Department of Computer Science \\
  Seoul Women's University \\
  Seoul 01797, Republic of Korea \\
  \texttt{hansanglee@swu.ac.kr} \\
  \And
  Haeil Lee \\
  LG Energy Solution \\
  Seoul 07335, Republic of Korea \\
  \texttt{haeil.lee@lgensol.com} \\
  \And
  Junmo Kim\thanks{Corresponding author.} \\
  School of Electrical Engineering \\
  KAIST \\
  Daejeon 34141, Republic of Korea \\
  \texttt{junmo.kim@kaist.ac.kr} \\
}

\begin{document}

\maketitle

\input{sec_0_abstract}

\input{sec_1_introduction}
\input{sec_2_relatedworks}
\input{sec_3_methods}
\input{sec_4_experiments}
\input{sec_5_results}
\input{sec_6_discussion}
\input{sec_7_conclusion}

\begin{ack}
This work was supported by the Center for Applied Research in Artificial Intelligence (CARAI) grant funded by Defense Acquisition Program Administration (DAPA) and Agency for Defense Development (ADD) (UD230017TD), and by the National Research Foundation of Korea (NRF) grant funded by the Korea government (MSIT) (RS-2025-00520184).
\end{ack}







\bibliographystyle{plainnat}
\bibliography{main}







\end{document}

%% file: sec_0_abstract.tex
\begin{abstract}
Modern vision models have achieved strong object-recognition performance, yet it remains unclear whether their representations encode object-level semantic relatedness, the meaningful connection between distinct object concepts that supports human visual cognition. Existing benchmarks predominantly target category prediction or rely on image--text matching, leaving the structure of the visual representation itself underexamined. Drawing on cognitive psychology, we recast semantic relatedness as a triplet-ranking task and study two image-only test beds: POPORO, an existing 400-triplet psychological stimulus set repurposed for representation evaluation, and PoporoIN, a newly constructed and manually curated 1{,}000-triplet ImageNet-validation extension. Each triplet is annotated along two orthogonal axes: a related-target axis distinguishing Categorical Relatedness (CR, taxonomic) from conTextual Relatedness (TR, thematic), and a distractor axis distinguishing Color-matched Distractors (CD) from Shape-matched Distractors (SD). Twenty pretrained models spanning supervised, self-supervised, vision--language, and generative paradigms were evaluated by cosine similarity in a non-training, inference-only protocol. Under this image-only triplet-ranking protocol, transformer-based representations exceeded convolutional counterparts by up to 18.30 percentage points on PoporoIN at comparable ImageNet accuracy, and vision--language pretrained encoders exceeded vision-only counterparts by up to 22.50 percentage points under matched ImageNet accuracy on POPORO. Across paradigms, models recognized taxonomic targets more reliably than thematic ones and were more easily misled by shape-matched than by color-matched distractors. The benchmarks expose representational properties that classification accuracy alone does not fully predict, providing a bridge between cognitive psychology and visual representation evaluation.
\end{abstract}
 

%% file: sec_1_introduction.tex
\begin{figure*}[!t]
\centering
\includegraphics[width=\textwidth]{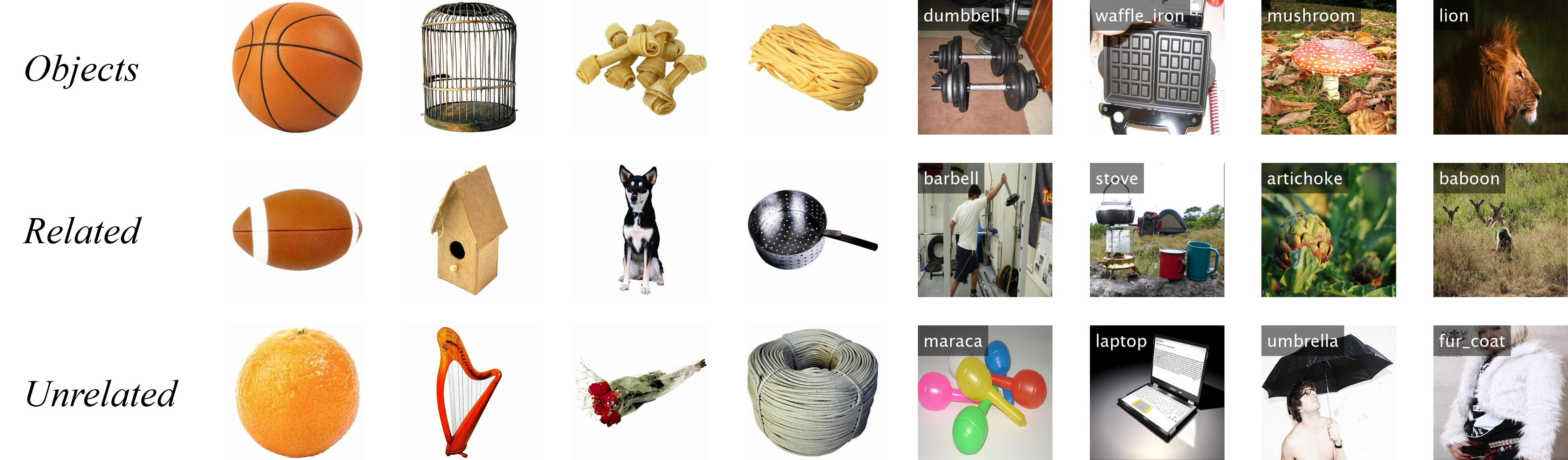}
\caption{\textbf{Object-level semantic relatedness as an image-only triplet-ranking task.} Each column shows one trial: a query \emph{object} (top), a semantically \emph{related} target (middle), and a semantically \emph{unrelated} distractor (bottom). The unrelated images deliberately share salient color or shape with the object so that purely perceptual similarity cannot solve the task. A vision model passes the trial when it places the related image closer to the object than the distractor in feature space. The eight columns span both POPORO (left, isolated objects on neutral background) and PoporoIN (right, ImageNet validation images with class labels) and cover all four cells of the annotation scheme (CR, TR, CD, SD).}
\label{fig:teaser}
\end{figure*}

\section{Introduction}
Visual cognition combines two abilities: identifying what an object is and inferring how it relates to other objects in meaning. For artificial systems, the second ability would support tasks ranging from scene understanding to grounded reasoning, where the relations among objects, rather than their identities alone, often determine the interpretation of an image. Cognitive psychology classifies the meaningful connections between object concepts under the umbrella of \emph{semantic relatedness}, which encompasses both \emph{taxonomic relations} arising from shared category membership (e.g., dog--wolf) and \emph{thematic relations} arising from functional or contextual co-occurrence (e.g., dog--leash) \cite{kacmajor2019,estes2011}. Whether modern vision models internalize this object-level structure in their visual representations remains an open empirical question.
 
Most benchmarks evaluating learned visual representations target category recognition. ImageNet \cite{deng2009imagenet} and COCO \cite{lin2014coco} assess single-object identification or detection, while Visual Relationship Detection (VRD) \cite{lu2016vrd} and Visual Genome \cite{krishna2017visualgenome} capture spatial or predicate relations within a single image. More recently, vision--language model (VLM) compositionality benchmarks, including ARO \cite{yuksekgonul2023aro}, SugarCrepe \cite{hsieh2023sugarcrepe}, Winoground \cite{thrush2022winoground}, MMVP \cite{tong2024mmvp}, and MMComposition \cite{hua2024mmcomposition}, probe sensitivity to attributes, relations, and word order. They do so through image--text matching, leaving the visual representation itself measured only indirectly. None of these benchmarks isolates whether two object concepts are placed close together in an image-only representation space because of the meaning that links them.
 
Cognitive psychology has been a productive lens for inspecting deep representations. Ritter et al.\ \cite{ritter2017cogpsych} showed that ImageNet-trained networks exhibit a shape bias, and subsequent work on texture bias \cite{geirhos2019,hermann2020}, convolutional--transformer comparisons \cite{tuli2021}, and shape bias in VLMs \cite{gavrikov2024} has refined this picture. Within cognitive science itself, the distinction between taxonomic and thematic relations is now a standard organizing axis of conceptual memory research \cite{estes2011,mirman2017}. Despite this rich foundation, no widely adopted vision benchmark explicitly distinguishes taxonomic from thematic targets while also separating perceptual confounds (color and shape) on the distractor side, leaving cognitive structure underused as a probe of visual representations.
 
The combination we identify, namely an inter-image, isolated-object, abstract-semantic, and image-only evaluation, has, to our knowledge, few direct counterparts in the existing literature. VRD and scene graph generation address within-image relations \cite{lu2016vrd,krishna2017visualgenome,xu2017scenegraph}; spatial reasoning benchmarks such as WhatsUp \cite{kamath2023whatsup} and VSR \cite{liu2023vsr} target spatial predicates; compositionality benchmarks rely on captions; and concept-association studies \cite{yamada2022cab} expose VLM failure modes rather than positively quantifying which pairs a representation places closer than chance. As a result, it remains unknown whether stronger object recognition translates into a representation that mirrors human-defined relatedness, and which architectural or training-paradigm choices govern that translation.

Fig.~\ref{fig:teaser} previews the task at a glance. To address the gap above, we adopt a triplet-ranking protocol grounded in cognitive psychology. We adopt POPORO \cite{kovalenko2012poporo}, a 400-triplet stimulus set originally collected for human visual relatedness judgments, as a psychologically validated test bed for our protocol, and we construct PoporoIN, a 1,000-triplet extension built from ImageNet validation images and curated through four large language model (LLM) prompts paired with manual review. Each triplet is annotated along two orthogonal axes: a related-target axis distinguishing Categorical Relatedness (CR, taxonomic) from conTextual Relatedness (TR, thematic), and a distractor axis distinguishing Color-matched Distractors (CD) from Shape-matched Distractors (SD). Twenty pretrained models spanning supervised, self-supervised, vision--language, and generative paradigms were evaluated by cosine similarity, without any task-specific training.
 
The benchmarks reveal patterns that classification accuracy alone does not fully predict: transformer- and vision--language-pretrained encoders consistently outperform convolutional and vision-only counterparts under matched ImageNet accuracy, and asymmetries along the taxonomic--thematic and color--shape axes are visible across paradigms. Detailed cross-paradigm patterns are reported in Section~\ref{sec:results}.
 
The main contributions of this work are as follows:
\begin{itemize}
    \item We introduce an image-only triplet-ranking protocol for evaluating whether pretrained visual representations encode object-level semantic relatedness without recourse to text or task-specific training, repurposing the existing POPORO stimulus set \cite{kovalenko2012poporo} as a psychologically validated test bed and constructing PoporoIN, a 1{,}000-triplet ImageNet-based extension that broadens the visual distribution and balances the four annotation cells.
    \item We translate the cognitive-psychology distinction between taxonomic and thematic relations into a 4-subgroup annotation scheme (CR, TR, CD, SD) organized along two orthogonal axes, enabling diagnostic separation of related-target and distractor effects.
    \item We define two diagnostic gap metrics, $\Delta_{\text{CR-TR}}$ and $\Delta_{\text{CD-SD}}$, that summarize where in representational space a given model is weak.
    \item We provide a cross-paradigm evaluation of twenty pretrained visual encoders and use the diagnostic gaps to summarize architecture-, pretraining-, and confound-related patterns in object-level relatedness encoding.
    \item We document that ImageNet classification accuracy alone does not fully predict object-level semantic relatedness, identifying a representational property that benchmarks targeting recognition do not characterize.
\end{itemize}

%% file: sec_2_relatedworks.tex
\section{Related Work}\label{sec:related-work}
 
\subsection{Vision Benchmarks for Object Understanding}
ImageNet \cite{deng2009imagenet}, COCO \cite{lin2014coco}, Visual Genome \cite{krishna2017visualgenome}, and VRD \cite{lu2016vrd} have driven progress on recognition, detection, and within-image predicate relations. Scene graph generation \cite{xu2017scenegraph} extracts $\langle$subject, predicate, object$\rangle$ tuples from a single image. These benchmarks evaluate object identification or intra-image relational structure rather than the holistic semantic relatedness between two object concepts depicted in separate images, which is the dimension our benchmarks target.
 
Two consequences follow from this restriction. The first is methodological: a representation can be strong on category prediction yet say nothing about whether it places dog closer to wolf than to school bus, because category prediction does not require any inter-class geometry beyond linear separability. The second is conceptual: when relatedness is reported at all in the existing benchmarks, it is reported through within-image predicate labels rather than through the geometry of the visual representation, leaving the abstract semantic relations between object concepts unmeasured. Our triplet-ranking protocol is designed to address this gap rather than to compete with the recognition benchmarks.
 
\subsection{Compositionality and Relation Benchmarks for Vision--Language Models}
A recent line of work probes VLMs through image--text matching with controlled hard negatives. ARO \cite{yuksekgonul2023aro} demonstrates that contrastively trained VLMs behave in a bag-of-words manner; SugarCrepe \cite{hsieh2023sugarcrepe} addresses the hard-negative bias of ARO using fluent LLM-generated negatives; Winoground \cite{thrush2022winoground} introduces minimally contrastive image--caption pairs on which CLIP performs near chance; and MMVP \cite{tong2024mmvp}, MMComposition \cite{hua2024mmcomposition}, VL-CheckList \cite{zhao2022vlchecklist}, and WhatsUp \cite{kamath2023whatsup} extend this style of probing to fine-grained composition and spatial predicates. All these benchmarks operate through caption manipulation. Our protocol differs in two ways: it removes text from the inference loop entirely and it ranks the relatedness of two object images against the same query image, evaluating whether the visual representation by itself encodes object-level relatedness. The Concept Association Bias study \cite{yamada2022cab} also focuses on VLMs but diagnoses a failure mode (bag-of-concepts behavior); our protocol instead provides a positive measurement of how reliably a representation aligns with human-defined relatedness.
 
\subsection{Cognitive Psychology Approaches in Computer Vision}
Cognitive psychology has informed vision research primarily through bias diagnostics. Ritter et al.\ \cite{ritter2017cogpsych} adapted shape-bias paradigms to deep networks; Geirhos et al.\ \cite{geirhos2019,gavrikov2024} and Hermann et al.\ \cite{hermann2020} characterized texture bias in convolutional networks and traced its origins, while Tuli et al.\ \cite{tuli2021} compared shape bias between convolutional networks and Vision Transformers (ViTs). The taxonomic--thematic distinction \cite{estes2011,mirman2017,kacmajor2019} has, in contrast, seen limited adoption in vision evaluation. Our benchmarks import this distinction directly into the related-target axis of the triplet design, complementing existing perceptual-bias studies with a representational-geometry probe of high-level conceptual structure.
 
The shape-bias literature and the present work measure related but distinct phenomena. Shape bias is a property of \emph{classification} under cue conflict, where higher reliance on shape is taken as more human-aligned. The SD axis of our benchmarks measures \emph{ranking} susceptibility, where shape similarity acts as a confound that misleads the representation into reporting semantic relatedness. A representation can simultaneously be human-aligned in cue-conflict classification and be misled by shape-matched distractors in triplet ranking; the two are not contradictory. This conceptual separation is one of the reasons we treat CR/TR and CD/SD as orthogonal annotation axes rather than collapsing them into a single four-class label.
 
\subsection{Representation Learning and Feature Quality}
Supervised \cite{he2016resnet,dosovitskiy2021vit,tan2019efficientnet}, self-supervised \cite{caron2020swav,caron2021dino,oquab2023dinov2,bardes2022vicreg}, vision--language \cite{radford2021clip,schuhmann2022laion}, and generative \cite{dhariwal2021guided,donahue2019bigbigan} pretraining have produced representations whose quality is typically assessed via downstream classification accuracy, linear probing, or k-nearest-neighbor retrieval. These protocols measure how separable the categories are. Our benchmarks measure a complementary property: whether the geometry of the representation reflects the relatedness of object concepts. This casts representation quality not as classification readiness but as alignment with human-defined conceptual structure.

\begin{figure*}[!t]
\centering
\includegraphics[width=\textwidth]{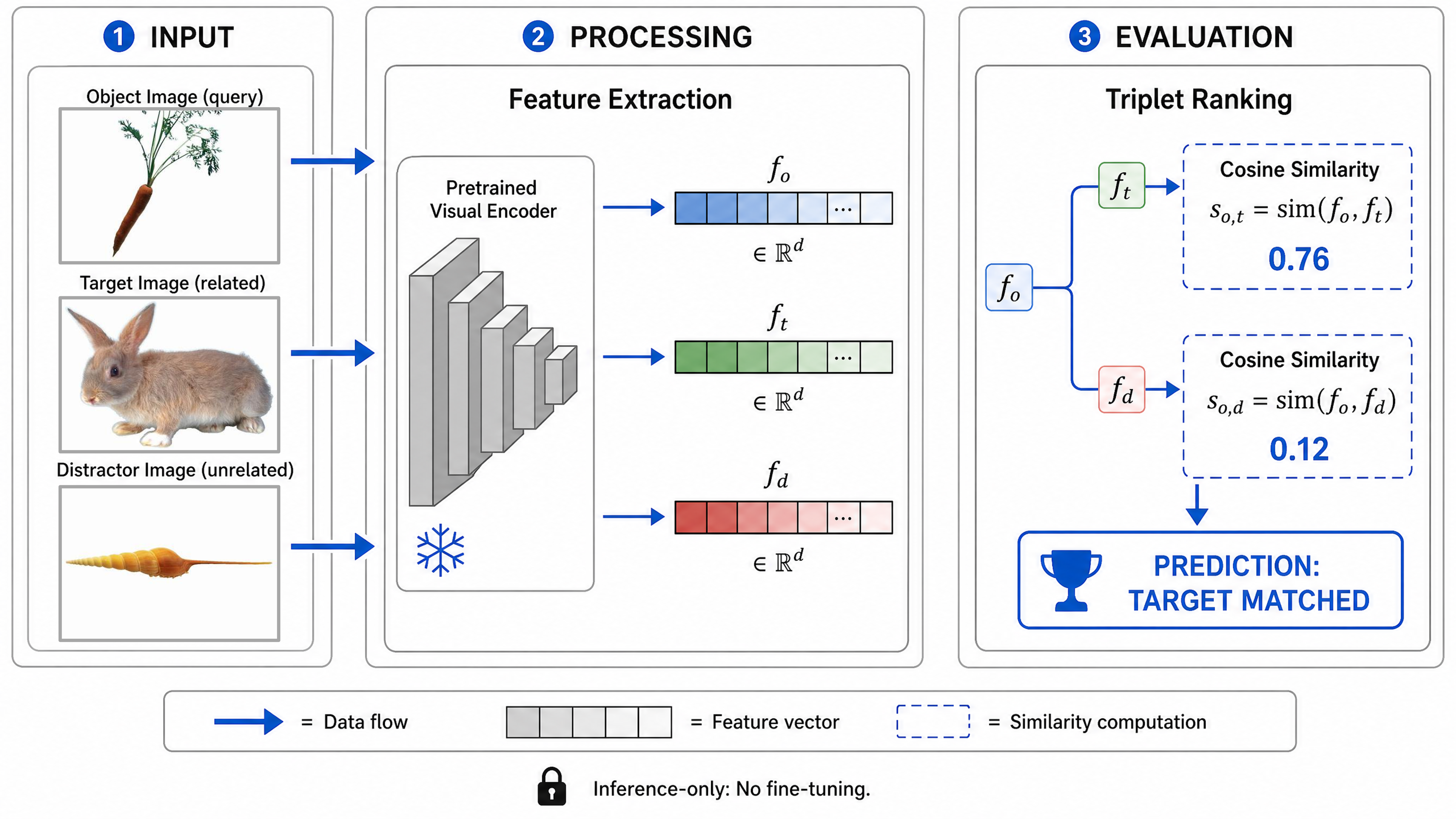}
\caption{\textbf{Image-only triplet-ranking pipeline.} A frozen pretrained visual encoder extracts feature vectors $f_o, f_t, f_d \in \mathbb{R}^{d}$ from the object image, the related target, and the unrelated distractor. Cosine similarities $s_{o,t}$ and $s_{o,d}$ are then computed between the object feature and each candidate feature, and the candidate with the higher similarity is selected as the model's predicted match. The encoder is never fine-tuned on the benchmark triplets.}
\label{fig:pipeline}
\end{figure*}
 
\subsection{Distinguishing Object-Level Semantic Relatedness from Adjacent Concepts}\label{sec:positioning}
Because the term ``relation'' is overloaded in vision, we briefly state what the present study does and does not measure. (i) Object/visual relationship detection and scene graph generation \cite{lu2016vrd,krishna2017visualgenome,xu2017scenegraph} model \emph{within-image} predicate structure, whereas we compare two object concepts depicted in \emph{separate} images. (ii) Spatial reasoning benchmarks \cite{kamath2023whatsup,liu2023vsr} target spatial predicates such as left/right or above/below; our protocol does not use spatial information. (iii) VLM compositionality benchmarks \cite{yuksekgonul2023aro,hsieh2023sugarcrepe,thrush2022winoground,tong2024mmvp} rely on image--text matching; ours is image-only triplet ranking. (iv) Concept association bias \cite{yamada2022cab} diagnoses a malfunction of binding; we positively quantify how reliably representations encode relatedness. The four properties --- inter-image, isolated-object, abstract-semantic, and image-only --- jointly define the evaluation space we target.

%% file: sec_3_methods.tex
\section{Benchmark and Methodology}\label{sec:method}

Our evaluation has three components: a definition of object-level semantic relatedness with the four-cell annotation scheme, two image-only triplet-ranking test beds---POPORO, an existing psychological stimulus set repurposed as a test bed, and PoporoIN, a newly constructed and curated ImageNet-based benchmark extension---and an inference-only matching protocol on top of frozen pretrained features. Section~\ref{sec:problem-formulation} fixes the terminology and the two-axis design; Sections~\ref{sec:poporo} and \ref{sec:poporoin} describe the two test beds; and Sections~\ref{sec:matching} and \ref{sec:gap-metrics} specify the matching protocol and the diagnostic metrics derived from per-subgroup accuracy. The full pipeline is illustrated in Fig.~\ref{fig:pipeline}. Across all components the design is deliberately minimal: no fine-tuning, no auxiliary training set, and a similarity function (cosine) that is invariant to feature scaling. This minimality places the question squarely on the pretrained representation rather than on the protocol that probes it.

\begin{table}[t!]
\centering
\caption{Two orthogonal annotation axes of the triplet design. Each triplet receives one label from the related-target axis (CR or TR) and one label from the distractor axis (CD or SD). Examples are illustrative of pairs that fall into each cell.}
\label{tab:taxonomy}
\begin{tabularx}{\linewidth}{l l >{\raggedright\arraybackslash}X}
\toprule
Axis & Label & Description and example pair \\
\midrule
\multirow{2}{*}{Related target}
  & CR (Categorical, taxonomic) & Same category or species, e.g., dog--wolf, microphone--loudspeaker \\
  & TR (conTextual, thematic)   & Functional, situational, or locational co-occurrence, e.g., dog--leash, corkscrew--wine bottle \\
\midrule
\multirow{2}{*}{Distractor}
  & CD (Color-matched) & Semantically unrelated, similar color, e.g., banana--school bus \\
  & SD (Shape-matched) & Semantically unrelated, similar shape, e.g., basketball--orange \\
\bottomrule
\end{tabularx}
\end{table}
 
\subsection{Object-Level Semantic Relatedness}\label{sec:problem-formulation}
Following the cognitive-psychology literature \cite{kacmajor2019,estes2011}, we define object-level semantic relatedness as the degree to which two object concepts are meaningfully connected, either through taxonomic/category membership or through contextual/thematic association. A triplet $(\mathbf{x}_o, \mathbf{x}_t, \mathbf{x}_d)$ consists of a query image $\mathbf{x}_o$ depicting an isolated object, a related target image $\mathbf{x}_t$, and an unrelated distractor image $\mathbf{x}_d$. A trial is counted as correct when the related target is closer to the query than the unrelated distractor in feature space.
 
Each triplet is annotated along two orthogonal axes, illustrated in Table~\ref{tab:taxonomy}. The related-target axis distinguishes Categorical Relatedness (CR), corresponding to taxonomic relations, from conTextual Relatedness (TR), corresponding to thematic relations. The distractor axis distinguishes Color-matched Distractors (CD) from Shape-matched Distractors (SD). CR and TR describe the type of semantically related target, whereas CD and SD describe the type of semantically unrelated distractor. They are not four mutually exclusive semantic relation classes but two orthogonal annotation axes of the triplet design, so each triplet receives one label from each axis; for example, the triplet \{banana, apple, school bus\} is simultaneously a CR (taxonomic) and a CD (color-matched) trial. 
 
We emphasize that SD is not equivalent to the shape-bias phenomenon studied in \cite{ritter2017cogpsych,geirhos2019,gavrikov2024}. Shape bias is measured by cue-conflict classification, where higher shape reliance is treated as more human-aligned. SD measures whether shape similarity \emph{misleads} a representation into reporting semantic relatedness in a triplet-ranking task. The two measurements are related but conceptually distinct. We accordingly use the verbs \emph{encode} and \emph{align with} rather than \emph{understand} or \emph{know} when describing model behavior, to keep claims at the level of representation geometry.

\begin{figure}[!t]
\centering
    \begin{subfigure}[b]{0.15\textwidth}
        \includegraphics[width=\textwidth]{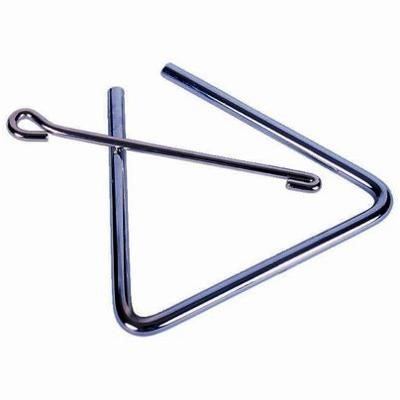}
    \end{subfigure}
    \begin{subfigure}[b]{0.15\textwidth}
        \includegraphics[width=\textwidth]{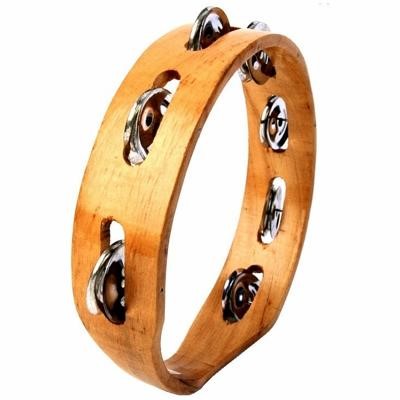}
    \end{subfigure}
    \begin{subfigure}[b]{0.15\textwidth}
        \includegraphics[width=\textwidth]{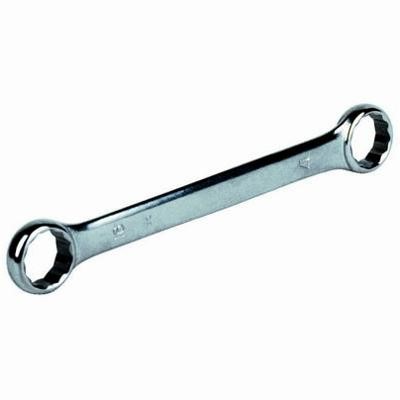}
    \end{subfigure}
    \\
    \begin{subfigure}[b]{0.15\textwidth}
        \includegraphics[width=\textwidth]{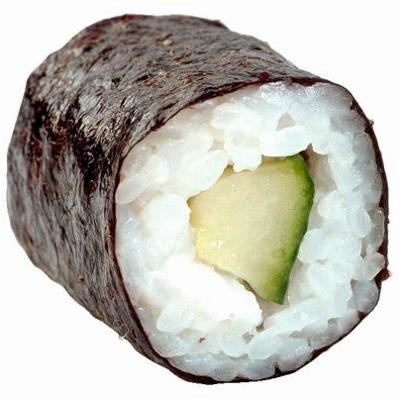}
    \end{subfigure}
    \begin{subfigure}[b]{0.15\textwidth}
        \includegraphics[width=\textwidth]{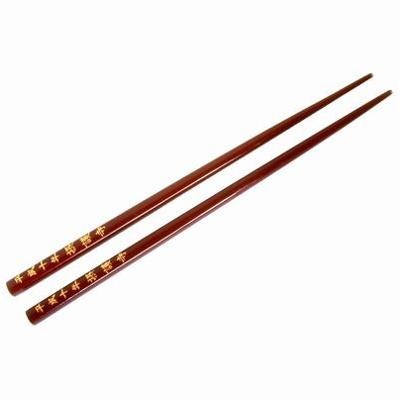}
    \end{subfigure}
    \begin{subfigure}[b]{0.15\textwidth}
        \includegraphics[width=\textwidth]{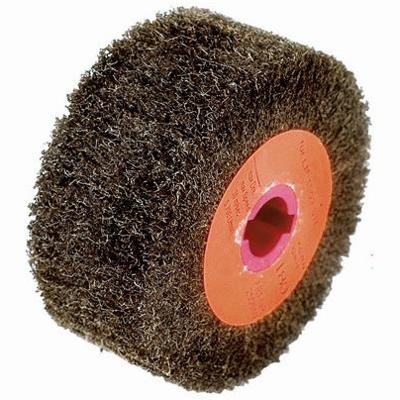}
    \end{subfigure}
    \\
    \begin{subfigure}[b]{0.15\textwidth}
        \includegraphics[width=\textwidth]{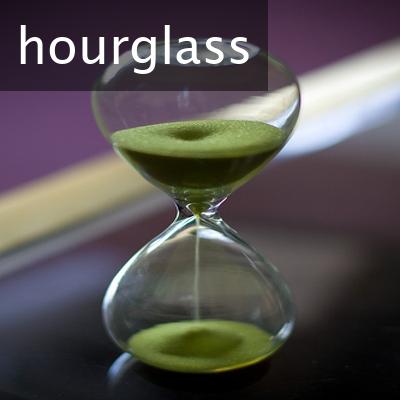}
    \end{subfigure}
    \begin{subfigure}[b]{0.15\textwidth}
        \includegraphics[width=\textwidth]{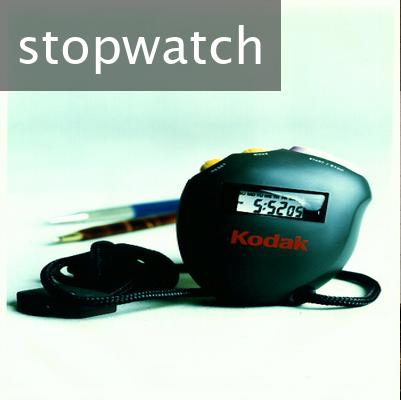}
    \end{subfigure}
    \begin{subfigure}[b]{0.15\textwidth}
        \includegraphics[width=\textwidth]{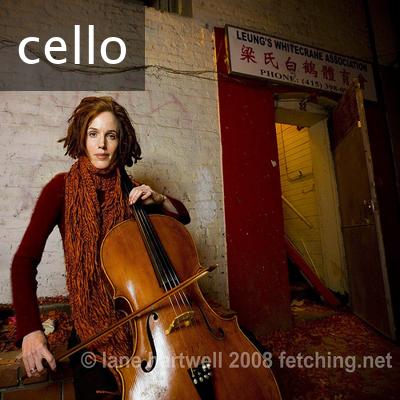}
    \end{subfigure}
    \\
    \begin{subfigure}[b]{0.15\textwidth}
        \includegraphics[width=\textwidth]{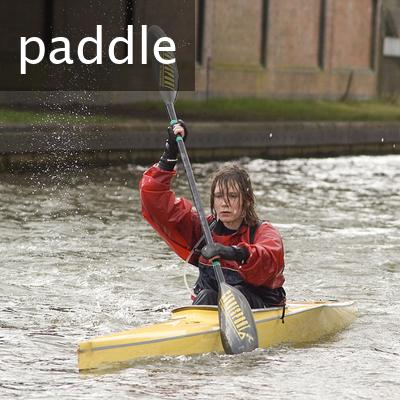}
        \caption{Objects}
    \end{subfigure}
    \begin{subfigure}[b]{0.15\textwidth}
        \includegraphics[width=\textwidth]{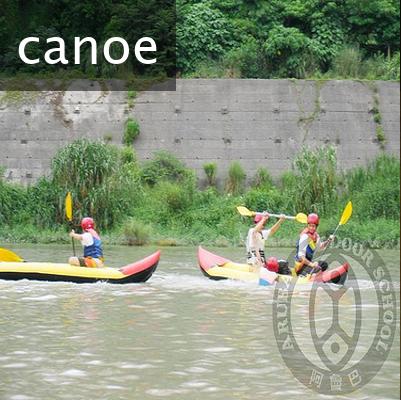}
        \caption{Related}
    \end{subfigure}
    \begin{subfigure}[b]{0.15\textwidth}
        \includegraphics[width=\textwidth]{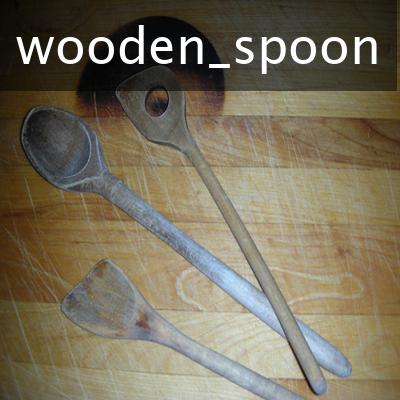}
        \caption{Unrelated}
    \end{subfigure}
\caption{\textbf{Representative triplets from the two benchmarks.} (Top two rows) POPORO: isolated-object stimuli on neutral backgrounds, originally curated for human relatedness studies. (Bottom two rows) PoporoIN: ImageNet validation images overlaid with their class labels. Each row contains one triplet --- query object (a), semantically related target (b), unrelated distractor (c) --- and the displayed triplets jointly cover all four cells of the annotation scheme in Table~\ref{tab:taxonomy}.}
\label{fig:examples}
\end{figure}
 
\subsection{POPORO Benchmark}\label{sec:poporo}
POPORO (Pool of Pairs of Related Objects) \cite{kovalenko2012poporo} is a stimulus set originally curated by cognitive psychologists for studying the semantic and contextual processing of visual stimuli. It consists of 400 triplets, each containing a query image, a related image, and an unrelated image; semantic relatedness was rated by 132 human participants. Within the related axis, 215 triplets are CR and 185 are TR; within the distractor axis, 170 are CD and 230 are SD. Because each triplet is labeled on both axes, the overall accuracy equals the share of correctly matched triplets on either axis. POPORO is small but psychologically validated and serves as the primary inference-only test bed in our experiments.
 
We treat POPORO strictly as a non-training test set, never as a source of training or fine-tuning signal. The motivation is that the benchmark question is whether pretraining alone produces representations that align with human-defined relatedness; introducing a training stage on POPORO triplets would conflate representation quality with adaptation. The original POPORO publication \cite{kovalenko2012poporo} reports human relatedness ratings rather than triplet-ranking accuracy under our cosine-similarity protocol, so a directly comparable human baseline is not available.

\begin{table}[!t]
\caption{The four LLM prompts used for PoporoIN pair generation, one per annotation cell. Each prompt was issued to ChatGPT (GPT-4o) and Gemini (Gemini 1.5 Pro) in parallel, paired with the ImageNet class list. Prompt P2 captures TR pairs across two complementary subtypes (functional/situational co-occurrence and negative-definition object--location or agent--instrument relations) within a single query.}
\label{tab:prompts}
\centering
\small
\begin{tabular}{c l p{0.55\linewidth}}
\toprule
ID & Target cell & Prompt text \\
\midrule
P1 & CR & From the ImageNet class list, list pairs of classes that share the same basic-level category or the same purpose (e.g., \emph{dog--wolf}, \emph{microphone--loudspeaker}). \\
P2 & TR & From the ImageNet class list, list pairs of classes that are thematically related rather than taxonomically related. The pair should fall into one of two subtypes: (i)~functionally complementary or routinely co-occurring in the same situation (e.g., \emph{corkscrew--wine bottle}, \emph{drum--drumsticks}), or (ii)~meaningfully related while explicitly \emph{not} belonging to the same species or basic-level category, such as object--location or agent--instrument relations (e.g., \emph{fish--aquarium}, \emph{musician--violin}). \\
P3 & SD & From the ImageNet class list, list pairs of classes that share a salient overall shape but are semantically unrelated (e.g., \emph{basketball--orange}). \\
P4 & CD & From the ImageNet class list, list pairs of classes that share a salient color but are semantically unrelated (e.g., \emph{banana--school bus}). \\
\bottomrule
\end{tabular}
\end{table}
 
\subsection{PoporoIN Benchmark}\label{sec:poporoin}
To mitigate the limited size of POPORO and to align stimuli with the visual distribution that pretrained models have seen, we constructed PoporoIN, a 1,000-triplet extension built from ImageNet validation images. Construction proceeds in three stages. First, we presented two large language models (ChatGPT-4o, and Gemini 1.5 Pro) with the ImageNet class list and queried them in parallel with four prompts, one targeting each of the four annotation cells of our scheme. Prompt P1 elicits CR pairs (same purpose or same basic category); P2 elicits TR pairs along two complementary subtypes within a single prompt --- typical functional and situational thematic relations (used together, mutually complementary, occurring in the same situation) and \emph{negative-definition} thematic relations (related but explicitly not in the same species or categorical hierarchy, such as object--location or agent--instrument); P3 elicits SD pairs (similar shape, semantically unrelated); and P4 elicits CD pairs (similar color, semantically unrelated). The full prompt text used for each of the four prompts is provided in Table~\ref{tab:prompts}.
 
Second, the candidate pair lists were manually reviewed by the authors. Specifically, we removed pairs when (i) the relation depended primarily on cultural convention without a direct visual cue, (ii) the ImageNet class label was too broad to consistently support the intended relation, or (iii) the proposed distractor was semantically related to the query rather than merely perceptually similar. Two authors performed independent first-pass review and resolved disagreements by discussion.
Third, ImageNet validation images for the retained class pairs were sampled to assemble triplets. PoporoIN contains 1,000 triplets in total, evenly distributed across the four annotation cells (CR=500, TR=500, CD=500, SD=500), giving a more balanced evaluation than POPORO. 
 
Fig.~\ref{fig:examples} contrasts representative triplets from the two benchmarks. POPORO triplets are isolated-object cutouts on neutral backgrounds, mirroring the controlled stimulus design of the original psychological study \cite{kovalenko2012poporo}, whereas PoporoIN triplets are natural-scene ImageNet validation images annotated with their class labels. The two benchmarks therefore probe the same triplet-ranking task across two complementary visual distributions: stimulus-controlled and ecologically natural.

\begin{table}[!t]
\caption{Summary of the evaluation datasets. Both serve as inference-only test sets.}
\label{tab:datasets}
\centering
\begin{tabular}{l c c c}
\toprule
Dataset & Triplets & Related axis & Distractor axis \\
& & (CR/TR) & (CD/SD) \\
\midrule
POPORO   & 400   & 215 / 185 & 170 / 230 \\
PoporoIN & 1{,}000 & 500 / 500 & 500 / 500 \\
\bottomrule
\end{tabular}
\end{table}
 
\subsection{Feature Extraction and Triplet Matching}\label{sec:matching}
For each model under evaluation, we extract a feature vector for the query, related target, and unrelated distractor images using the standard inference protocol of the corresponding implementation. The goal of this protocol is not to optimize each model's maximum attainable relatedness score, but to compare pretrained representations under a uniform, architecture-agnostic readout; we therefore apply a single cosine-similarity readout on the standard inference-time feature for each model. For supervised and CLIP fine-tuned models loaded through the timm library \cite{wightman2019timm}, features are taken from the activation immediately preceding the classification head. For self-supervised models (VICReg \cite{bardes2022vicreg}, SwAV \cite{caron2020swav}, DINOv1 \cite{caron2021dino}, DINOv2 \cite{oquab2023dinov2}) we use the encoder output of the official checkpoint. For CLIP \cite{radford2021clip,schuhmann2022laion} we use the visual encoder only; the text encoder is not used in the main experiments. For Guided Diffusion \cite{dhariwal2021guided} we use the activation of block 24 at noise timestep $t=90$, following the feature-extraction protocol of \cite{mukhopadhyay2023diffusion}. For BigBiGAN \cite{donahue2019bigbigan} we use the encoder output. Per-model layer choices, pooling strategies, and normalization conventions are listed in the supplementary material.

Using only the visual encoder of CLIP is a deliberate choice: the question we evaluate is whether vision--language pretraining produces visual representations that encode object-level relatedness, not whether language information at inference time can solve the task. Holding the inference protocol uniformly image-only across all paradigms keeps the comparison fair; introducing zero-shot text matching for CLIP alone would conflate representation quality with inference-time linguistic reasoning. Quantifying the contribution of language at inference time, by comparing visual-only ranking with zero-shot text-based ranking under matched conditions, is left to future work.
 
Given features $\mathbf{f}_o$, $\mathbf{f}_t$, $\mathbf{f}_d$ of the query, related target, and unrelated distractor, we compute cosine similarity
\begin{equation}
\operatorname{sim}(\mathbf{a},\mathbf{b}) = \frac{\mathbf{a}^{\top}\mathbf{b}}{\lVert \mathbf{a}\rVert\,\lVert \mathbf{b}\rVert},
\label{eq:cossim}
\end{equation}
and predict the related image as
\begin{equation}
\hat{i} = \arg\max_{i \in \{t,d\}} \operatorname{sim}(\mathbf{f}_o, \mathbf{f}_i).
\label{eq:argmax}
\end{equation}
Accuracy is the fraction of triplets for which $\hat{i}=t$. Because evaluation is inference-only and cosine similarity is invariant to feature scale, results are deterministic for a given model and the choice of random seed is irrelevant. Cosine similarity is the simplest dimension-invariant comparison function and is widely used in representation evaluation.

\begin{table*}[!t]
\caption{Summary of evaluated pretrained models. ImageNet top-1 accuracy is taken from the corresponding official sources.}
\label{tab:models}
\centering
\begin{tabular}{l l l}
\toprule
Paradigm & Backbone(s) & Pretraining \\
\midrule
Supervised        & ResNet-50/101, ViT-B/32, ViT-B/16 & ImageNet \\
SSL (vision-only) & VICReg/SwAV/DINOv1 RN-50; DINOv1 ViT-B/16, B/8 & ImageNet \\
SSL (vision-only) & DINOv2 ViT-B/14, ViT-L/14 & LVD-142M \\
Vision--language  & CLIP RN-50/101, ViT-B/16, ViT-L/14 & WIT-400M \\
Vision--language  & CLIP ViT-B/16, ViT-L/14@336, ViT-H/14 & LAION-2B \\
Generative        & Guided Diffusion, BigBiGAN & ImageNet \\
\bottomrule
\end{tabular}
\end{table*}
 
\subsection{Diagnostic Gap Metrics}\label{sec:gap-metrics}
To summarize structural weaknesses of a representation we define two gap metrics, derived directly from per-subgroup accuracy:
\begin{align}
\Delta_{\text{CR-TR}} &= \mathrm{Acc}_{\text{CR}} - \mathrm{Acc}_{\text{TR}}, \label{eq:dcrtr}\\
\Delta_{\text{CD-SD}} &= \mathrm{Acc}_{\text{CD}} - \mathrm{Acc}_{\text{SD}}. \label{eq:dcdsd}
\end{align}
A positive $\Delta_{\text{CR-TR}}$ indicates that the representation recognizes taxonomic relatedness more reliably than thematic relatedness, paralleling the human cognitive pattern reported in \cite{estes2011,mirman2017}. A positive $\Delta_{\text{CD-SD}}$ indicates that shape-matched distractors mislead the representation more than color-matched ones. Both metrics are computed from the same accuracy table without additional experiments and serve as the diagnostic axis along which we summarize results.

%% file: sec_4_experiments.tex
\section{Experimental Setup}\label{sec:setup}
 
\subsection{Datasets}
We evaluate on POPORO (400 triplets; CR=215, TR=185, CD=170, SD=230) and PoporoIN (1,000 triplets; CR=500, TR=500, CD=500, SD=500). Both serve as inference-only test sets; no training or fine-tuning is performed on either benchmark. PoporoIN images are drawn from the ImageNet validation set. A summary is provided in Table~\ref{tab:datasets}.
 
\subsection{Models}
We evaluate twenty pretrained models, summarized in Table~\ref{tab:models}, spanning four paradigms. Supervised models (4): ResNet-50, ResNet-101, ViT-B/32, and ViT-B/16, all trained on ImageNet \cite{he2016resnet,dosovitskiy2021vit}. Self-supervised vision-only models (7): VICReg-RN50 \cite{bardes2022vicreg}, SwAV-RN50 \cite{caron2020swav}, DINOv1 with ResNet-50, ViT-B/16 and ViT-B/8 \cite{caron2021dino}, and DINOv2 with ViT-B/14 and ViT-L/14 \cite{oquab2023dinov2}. Vision--language CLIP models (7): four OpenAI CLIP variants (ResNet-50, ResNet-101, ViT-B/16, ViT-L/14) trained on WIT-400M \cite{radford2021clip}, and three OpenCLIP variants (ViT-B/16, ViT-L/14@336px, ViT-H/14) trained on LAION-2B \cite{schuhmann2022laion}. Generative models (2): Guided Diffusion \cite{dhariwal2021guided} and BigBiGAN \cite{donahue2019bigbigan}. 
 
\subsection{Implementation Details}
All evaluations are inference-only on a server equipped with eight NVIDIA A100 GPUs. No model is trained or fine-tuned in this study. Supervised and CLIP fine-tuned models are loaded through timm \cite{wightman2019timm}; self-supervised, CLIP, Guided Diffusion, and BigBiGAN models use their official implementations. Each model uses the input resolution and normalization statistics specified by its official inference pipeline, including 224$\times$224 resizing for most models and 336$\times$336 for selected CLIP variants. Because the protocol involves only forward passes, training-related hyperparameters (batch size, optimizer, scheduler, learning rate, random seed) are not applicable to this study, and inference is deterministic for a given model.
 
\subsection{Evaluation Metrics}
We report overall accuracy on POPORO and PoporoIN as well as per-subgroup accuracy on CR, TR, CD, and SD. The diagnostic gap metrics $\Delta_{\text{CR-TR}}$ and $\Delta_{\text{CD-SD}}$, defined in Section~\ref{sec:gap-metrics}, summarize structural asymmetries within a model. ImageNet top-1 classification accuracy of each backbone is reported as a reference and is used qualitatively, not as a target metric, when comparing object recognition with relatedness encoding.

%% file: sec_5_results.tex
\section{Results}\label{sec:results}
 
\subsection{Main Performance Comparison}\label{sec:main-results}
Table~\ref{tab:poporo} summarizes evaluation on POPORO and Table~\ref{tab:poporoin} on PoporoIN. The best POPORO accuracy is 86.00\% obtained by CLIP LAION-2B ViT-L/14@336px, and the best PoporoIN accuracy is 77.90\% obtained by CLIP WIT-400M ViT-L/14. The lowest POPORO accuracies (43.75\%) are produced by VICReg-RN50 and DINOv1-RN50, both at or below the 50\% chance level, indicating that perceptual confounds (color and shape) actively mislead these representations on POPORO triplets.

\begin{table}[t!]
\centering
\caption{Evaluation on POPORO (400 triplets). All values are accuracies (\%); higher is better. ImageNet column reports each backbone's reference top-1 accuracy, and the Params (M) column reports the visual-encoder parameter count in millions. The last two columns report the diagnostic gap metrics $\Delta_{\text{CR-TR}}=\text{Acc}_{\text{CR}}-\text{Acc}_{\text{TR}}$ and $\Delta_{\text{CD-SD}}=\text{Acc}_{\text{CD}}-\text{Acc}_{\text{SD}}$; larger absolute values indicate stronger asymmetry rather than better performance. Best accuracy per column is in \textbf{bold}; $\Delta$ values are not bolded.}
\label{tab:poporo}
\setlength{\tabcolsep}{3pt}
\renewcommand{\arraystretch}{1.05}
\footnotesize
\begin{tabular}{ll|rr|r|rrrr|rr}   
\toprule
\multicolumn{2}{c}{Models} 
& \makecell{Params\\(M)}                    
& \makecell{ImageNet\\Acc.}                       
& \makecell{POPORO\\Acc.}                      
& \multicolumn{4}{c}{Subgroup Accuracies}      
& \multicolumn{2}{c}{Gap Metrics} \\           
\cmidrule(lr){1-2} \cmidrule(lr){6-9} \cmidrule(lr){10-11}
Learning paradigm & Backbone & & & 
& CR & TR & CD & SD 
& $\Delta_{\text{CR-TR}}$ & $\Delta_{\text{CD-SD}}$ \\
\midrule
Supervised        & ResNet-50            & 25.56  & 80.37 & 48.25 & 51.63 & 44.32 & 54.71 & 43.48 & $+7.31$  & $+11.23$ \\
Supervised        & ResNet-101           & 44.55  & 81.94 & 48.00 & 48.84 & 47.03 & 46.47 & 49.13 & $+1.81$  & $-2.66$ \\
Supervised        & ViT-B/32             & 88.22  & 74.90 & 57.00 & 66.05 & 46.49 & 61.18 & 53.91 & $+19.56$ & $+7.27$ \\
Supervised        & ViT-B/16             & 86.57  & 79.15 & 63.75 & 74.42 & 51.35 & 67.06 & 61.30 & $+23.07$ & $+5.76$ \\
Self-Sup.\ (VICReg) & ResNet-50          & 25.56  & 73.20 & 43.75 & 49.30 & 37.30 & 50.59 & 38.70 & $+12.00$ & $+11.89$ \\
Self-Sup.\ (SwAV)   & ResNet-50          & 25.56  & 75.30 & 44.00 & 51.63 & 35.14 & 47.06 & 41.74 & $+16.49$ & $+5.32$ \\
Self-Sup.\ (DINOv1) & ResNet-50          & 25.56  & 75.30 & 43.75 & 49.77 & 36.76 & 47.06 & 41.30 & $+13.01$ & $+5.76$ \\
Self-Sup.\ (DINOv1) & ViT-B/16           & 86.57  & 78.20 & 50.50 & 61.86 & 37.30 & 54.71 & 47.39 & $+24.56$ & $+7.32$ \\
Self-Sup.\ (DINOv1) & ViT-B/8            & 86.58  & 80.10 & 56.75 & 67.44 & 44.32 & 62.35 & 52.61 & $+23.12$ & $+9.74$ \\
Self-Sup.\ (DINOv2) & ViT-B/14           & 86.00  & 84.50 & 69.50 & 76.28 & 61.62 & 80.00 & 61.74 & $+14.66$ & $+18.26$ \\
Self-Sup.\ (DINOv2) & ViT-L/14           & 300.0  & 86.30 & 74.75 & 80.00 & 68.65 & 75.88 & 73.91 & $+11.35$ & $+1.97$ \\
CLIP (WIT-400M)   & ResNet-50            & 25.56  & 73.30 & 66.25 & 74.88 & 56.22 & 70.00 & 63.48 & $+18.66$ & $+6.52$ \\
CLIP (WIT-400M)   & ResNet-101           & 44.55  & 75.70 & 80.75 & 86.51 & 74.05 & 84.71 & 77.83 & $+12.46$ & $+6.88$ \\
CLIP (WIT-400M)   & ViT-B/16             & 86.57  & 85.93 & 74.00 & 84.19 & 62.16 & 77.65 & 71.30 & $+22.03$ & $+6.35$ \\
CLIP (WIT-400M)   & ViT-L/14             & 304.2  & 88.17 & 84.75 & 90.70 & 77.84 & \textbf{88.82} & 81.74 & $+12.86$ & $+7.08$ \\
CLIP (LAION-2B)   & ViT-B/16             & 86.57  & 86.17 & 79.00 & 86.51 & 70.27 & 82.35 & 76.52 & $+16.24$ & $+5.83$ \\
CLIP (LAION-2B)   & ViT-L/14@336         & 304.5  & \textbf{88.18} & \textbf{86.00} & \textbf{92.09} & \textbf{78.92} & 87.06 & \textbf{85.22} & $+13.17$ & $+1.84$ \\
CLIP (LAION-2B)   & ViT-H/14             & 632.0  & 87.59 & 81.50 & 88.37 & 73.51 & 87.06 & 77.39 & $+14.86$ & $+9.67$ \\
Generative        & Guided Diffusion     & ${\approx}552$  & 61.95 & 49.00 & 51.63 & 45.95 & 55.29 & 44.35 & $+5.68$  & $+10.94$ \\
Generative        & BigBiGAN             & ${\approx}25.6$ & 60.80 & 47.25 & 53.02 & 40.54 & 52.94 & 43.04 & $+12.48$ & $+9.90$ \\
\bottomrule
\end{tabular}
\end{table}

\begin{table}[t!]
\centering
\caption{Evaluation on PoporoIN (1{,}000 triplets). All values are accuracies (\%); higher is better. The Params (M) column reports the visual-encoder parameter count in millions. The last two columns report the diagnostic gap metrics $\Delta_{\text{CR-TR}}$ and $\Delta_{\text{CD-SD}}$ defined in Section~\ref{sec:gap-metrics}; larger absolute values indicate stronger asymmetry rather than better performance. Best accuracy per column is in \textbf{bold}; $\Delta$ values are not bolded.}
\label{tab:poporoin}
\setlength{\tabcolsep}{3pt}
\renewcommand{\arraystretch}{1.05}
\footnotesize
\begin{tabular}{ll|rr|r|rrrr|rr}   
\toprule
\multicolumn{2}{c}{Models} 
& \makecell{Params\\(M)}                    
& \makecell{ImageNet\\Acc.}                       
& \makecell{PoporoIN\\Acc.}                      
& \multicolumn{4}{c}{Subgroup Accuracies}      
& \multicolumn{2}{c}{Gap Metrics} \\           
\cmidrule(lr){1-2} \cmidrule(lr){6-9} \cmidrule(lr){10-11}
Learning paradigm & Backbone & & & 
& CR & TR & CD & SD 
& $\Delta_{\text{CR-TR}}$ & $\Delta_{\text{CD-SD}}$ \\
\midrule
Supervised        & ResNet-50            & 25.56  & 80.37 & 55.90 & 56.80 & 55.00 & 60.20 & 51.60 & $+1.80$  & $+8.60$ \\
Supervised        & ResNet-101           & 44.55  & 81.94 & 47.90 & 50.60 & 45.20 & 47.00 & 48.80 & $+5.40$  & $-1.80$ \\
Supervised        & ViT-B/32             & 88.22  & 74.90 & 70.80 & 72.00 & 69.60 & 76.60 & 65.00 & $+2.40$  & $+11.60$ \\
Supervised        & ViT-B/16             & 86.57  & 79.15 & 74.20 & 73.40 & 75.00 & 78.80 & 69.60 & $-1.60$  & $+9.20$ \\
Self-Sup.\ (VICReg) & ResNet-50          & 25.56  & 73.20 & 53.60 & 54.00 & 53.20 & 56.60 & 50.60 & $+0.80$  & $+6.00$ \\
Self-Sup.\ (SwAV)   & ResNet-50          & 25.56  & 75.30 & 51.50 & 52.60 & 50.40 & 52.60 & 50.40 & $+2.20$  & $+2.20$ \\
Self-Sup.\ (DINOv1) & ResNet-50          & 25.56  & 75.30 & 53.80 & 55.60 & 52.00 & 57.20 & 50.40 & $+3.60$  & $+6.80$ \\
Self-Sup.\ (DINOv1) & ViT-B/16           & 86.57  & 78.20 & 65.40 & 67.60 & 63.20 & 70.80 & 60.00 & $+4.40$  & $+10.80$ \\
Self-Sup.\ (DINOv1) & ViT-B/8            & 86.58  & 80.10 & 66.20 & 69.60 & 62.80 & 73.00 & 59.40 & $+6.80$  & $+13.60$ \\
Self-Sup.\ (DINOv2) & ViT-B/14           & 86.00  & 84.50 & 63.40 & 66.60 & 60.20 & 68.60 & 58.20 & $+6.40$  & $+10.40$ \\
Self-Sup.\ (DINOv2) & ViT-L/14           & 300.0  & 86.30 & 63.40 & 65.40 & 61.40 & 66.00 & 60.80 & $+4.00$  & $+5.20$ \\
CLIP (WIT-400M)   & ResNet-50            & 25.56  & 73.30 & 68.90 & 70.40 & 67.40 & 73.80 & 64.00 & $+3.00$  & $+9.80$ \\
CLIP (WIT-400M)   & ResNet-101           & 44.55  & 75.70 & 68.60 & 70.40 & 66.80 & 71.80 & 65.40 & $+3.60$  & $+6.40$ \\
CLIP (WIT-400M)   & ViT-B/16             & 86.57  & 85.93 & 73.30 & 71.60 & 75.00 & 77.40 & 69.20 & $-3.40$  & $+8.20$ \\
CLIP (WIT-400M)   & ViT-L/14             & 304.2  & 88.17 & \textbf{77.90} & \textbf{74.00} & \textbf{81.80} & \textbf{82.00} & \textbf{73.80} & $-7.80$ & $+8.20$ \\
CLIP (LAION-2B)   & ViT-B/16             & 86.57  & 86.17 & 74.10 & 72.40 & 75.80 & 78.20 & 70.00 & $-3.40$  & $+8.20$ \\
CLIP (LAION-2B)   & ViT-L/14@336         & 304.5  & \textbf{88.18} & 72.50 & 67.80 & 77.20 & 75.80 & 69.20 & $-9.40$ & $+6.60$ \\
CLIP (LAION-2B)   & ViT-H/14             & 632.0  & 87.59 & 70.60 & 71.60 & 69.60 & 74.60 & 66.60 & $+2.00$  & $+8.00$ \\
Generative        & Guided Diffusion     & ${\approx}552$  & 61.95 & 60.50 & 62.40 & 58.60 & 66.00 & 55.00 & $+3.80$  & $+11.00$ \\
Generative        & BigBiGAN             & ${\approx}25.6$ & 60.80 & 55.60 & 57.40 & 53.80 & 58.40 & 52.80 & $+3.60$  & $+5.60$ \\
\bottomrule
\end{tabular}
\end{table}

\begin{figure}[!t]
\centering
\includegraphics[width=0.6\columnwidth]{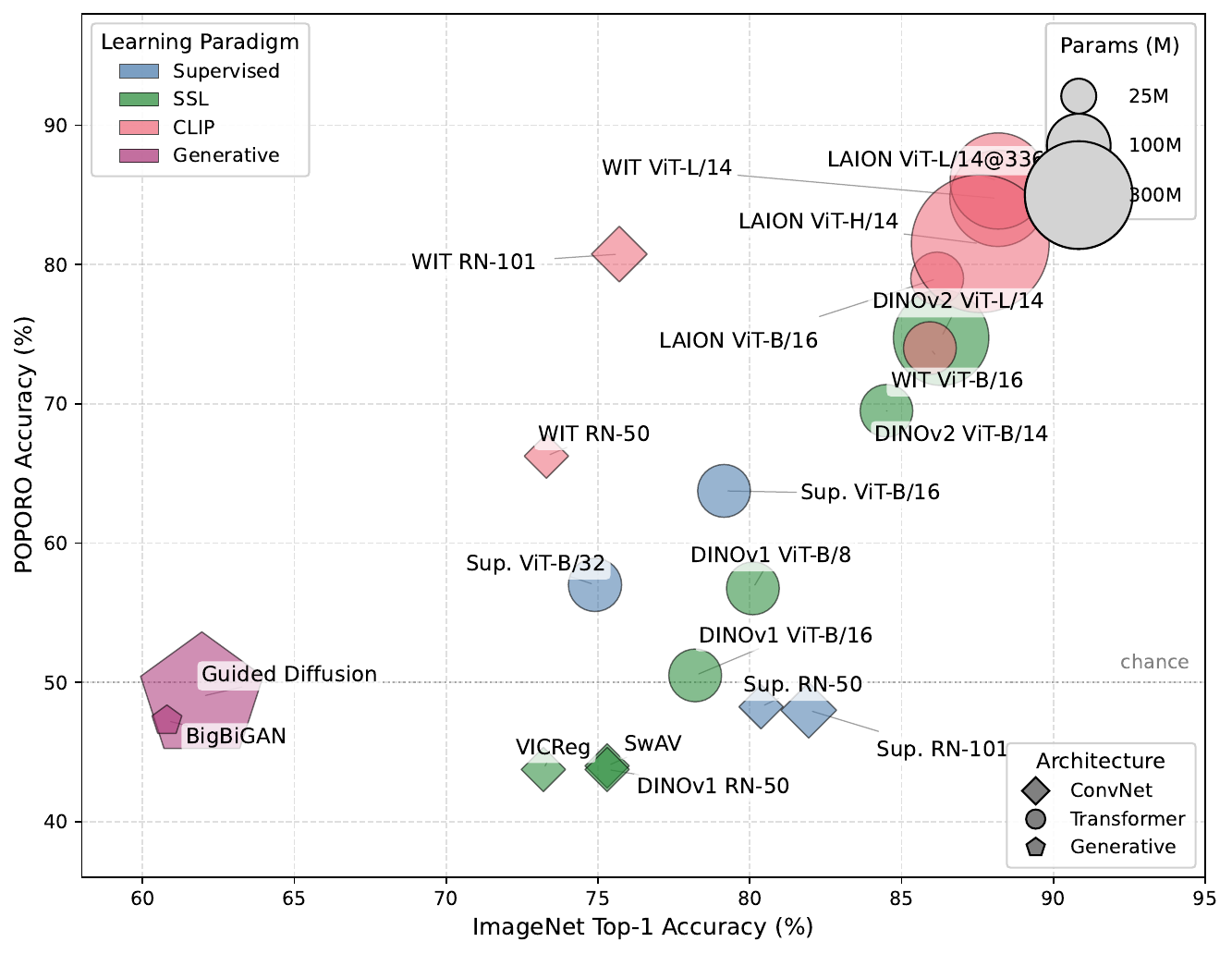}\\[0.3em]
\small (a) POPORO\\[0.6em]
\includegraphics[width=0.6\columnwidth]{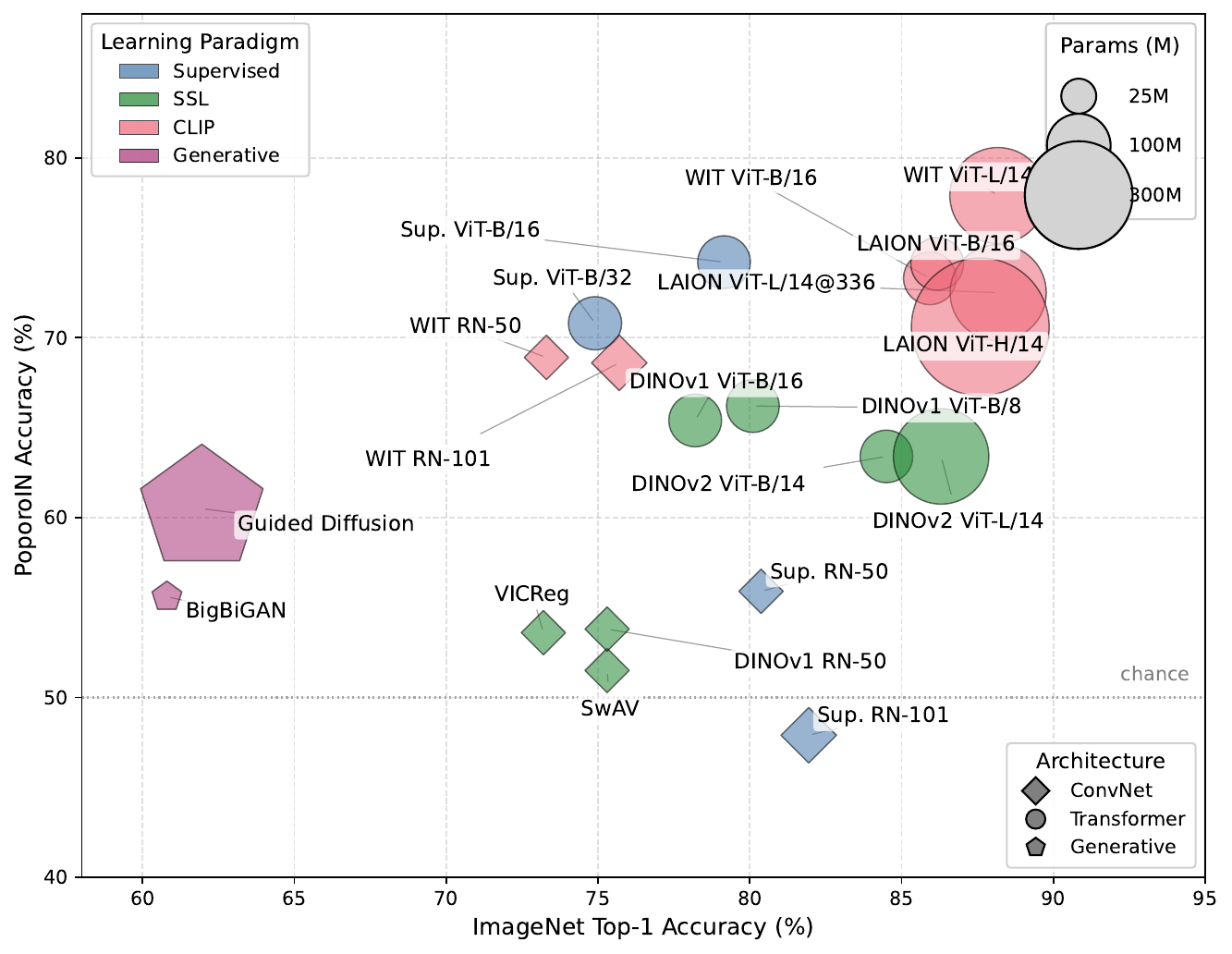}\\[0.3em]
\small (b) PoporoIN
\caption{\textbf{Object-level relatedness accuracy versus ImageNet accuracy.} Each marker represents one of the twenty evaluated models on (a) POPORO and (b) PoporoIN. Marker area is proportional to the visual-encoder parameter count, color encodes the learning paradigm, and marker shape encodes the architectural family (ConvNet, Transformer, or Generative). Paradigm color separates the points more cleanly than marker size does at comparable ImageNet accuracy: CLIP markers (pink) sit above SSL (green) and Supervised (blue) markers at matched ImageNet level on both panels.}
\label{fig:scatter}
\end{figure}
 
\subsection{Architectural Family: Transformer versus Convolutional Network}
At comparable ImageNet accuracy, transformer backbones place semantically related objects closer together than convolutional backbones. Comparing supervised ViT-B/16 (ImageNet 79.15\%) with supervised ResNet-50 (ImageNet 80.37\%), POPORO accuracy increases from 48.25\% to 63.75\% (+15.50 percentage points) and PoporoIN accuracy from 55.90\% to 74.20\% (+18.30 percentage points), despite a slightly lower ImageNet score for the ViT. The supervised ResNet pair (RN-50 $\to$ RN-101, ImageNet 80.37\% $\to$ 81.94\%) further illustrates that within-family ConvNet scaling does not lift POPORO accuracy out of the near-chance range (48.25\% $\to$ 48.00\%), while a single architectural switch to ViT at comparable ImageNet level produces a 15-percentage-point jump. The same architectural separation is visible in Fig.~\ref{fig:scatter}: at every horizontal slice of ImageNet accuracy, transformer (circle) markers sit above convolutional (diamond) markers of the same color, and the contrast is largest in the high-ImageNet region populated by ViT-L and ViT-H backbones. These results show that the evaluated transformer backbones achieve higher relatedness accuracy than convolutional backbones at comparable or higher ImageNet accuracy under the present protocol; mechanistic interpretations are deferred to Section~\ref{sec:discussion}.
 
\subsection{Learning Paradigm: Vision--Language versus Vision-Only}
At matched architecture and matched ImageNet accuracy, vision--language pretraining produces representations that better encode relatedness than vision-only pretraining. CLIP-RN50 (ImageNet 73.30\%) and VICReg-RN50 (ImageNet 73.20\%) share architecture and parameter count, yet the former reaches 66.25\% on POPORO and 68.90\% on PoporoIN, against 43.75\% and 53.60\% for the latter, a margin of +22.50 and +15.30 percentage points. At larger scale, CLIP WIT-400M ViT-L/14 outperforms DINOv2 ViT-L/14 by +10.00 percentage points on POPORO (84.75 vs.\ 74.75) and +14.50 on PoporoIN (77.90 vs.\ 63.40); part of this gap is attributable to the difference in pretraining-data scale (WIT-400M vs.\ LVD-142M), but the same-architecture, equal-ImageNet-accuracy RN-50 comparison is not explained by data scale alone. The paradigm separation is also visible in Fig.~\ref{fig:scatter}: CLIP markers (pink) sit above SSL (green) and Supervised (blue) markers at matched ImageNet accuracy across the full ImageNet-accuracy range, and the vertical separation is comparable in magnitude to the architectural separation reported in Section~V-B. We interpret the pattern as evidence that CLIP-style image--text pretraining is associated with stronger alignment to object-level relatedness under our protocol, while acknowledging that data scale, training objective, and architecture are not fully decoupled in this comparison.

\subsection{ImageNet Accuracy Does Not Fully Predict Relatedness}
Within a fixed architecture and learning paradigm, relatedness accuracy correlates with ImageNet accuracy, but the relationship breaks across paradigms. CLIP-RN50 (ImageNet 73.30\%) outperforms supervised ResNet-50 (ImageNet 80.37\%) by 18.00 percentage points on POPORO, and supervised ViT-B/16 (ImageNet 79.15\%) outperforms supervised ResNet-50 (ImageNet 80.37\%) by 15.50 percentage points despite a lower ImageNet score. Within-family ConvNet scaling is also weak: supervised ResNet-50 $\to$ ResNet-101 raises ImageNet accuracy from 80.37\% to 81.94\% but leaves POPORO accuracy essentially unchanged (48.25\% $\to$ 48.00\%). The same pattern is visible in Fig.~\ref{fig:scatter}: at fixed horizontal positions on the ImageNet axis, markers spread vertically by 20--30 percentage points on both panels, and the two generative representatives (pentagons) sit far above where their low ImageNet accuracy alone would place them, while several supervised ConvNets (blue diamonds) fall to or below chance despite high ImageNet scores. Object recognition accuracy is therefore insufficient on its own to characterize how a representation organizes object-level meaning.

\begin{figure*}[!t]
\centering
\includegraphics[width=\textwidth]{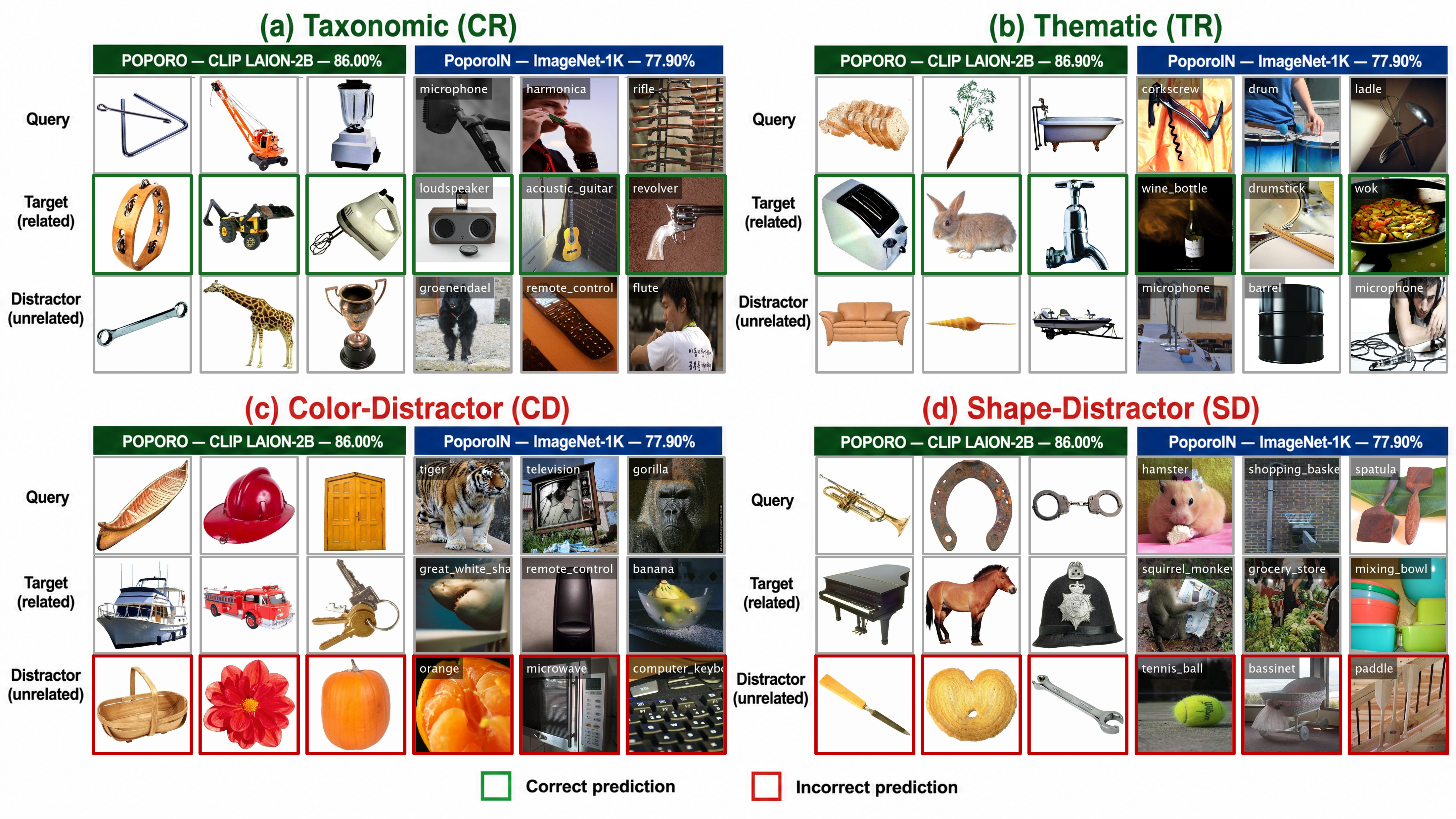}
\caption{\textbf{Qualitative agreement and disagreement cases of the strongest model on each benchmark.} Panels are arranged along the two orthogonal annotation axes of the triplet design. Top row --- agreements on the related-target axis: (a) CR (taxonomic) and (b) TR (thematic), where the model places the related target closer to the query than the distractor. Bottom row --- disagreements on the distractor axis: (c) CD (color-matched) and (d) SD (shape-matched), where the model is misled and places the perceptually similar but semantically unrelated distractor closer than the correct related target. Each panel shows three representative triplets, with each column displaying the query (top), the related target, and the distractor; PoporoIN cells overlay ImageNet class labels. Models illustrated are CLIP LAION-2B ViT-L/14@336px (best on POPORO, 86.00\%) and CLIP WIT-400M ViT-L/14 (best on PoporoIN, 77.90\%).}
\label{fig:qualitative}
\end{figure*}

\subsection{Model Size Does Not Account for the Architectural and Paradigm Patterns}\label{sec:model-size}
A natural concern about the architectural and paradigm patterns reported in Sections~V-B and V-C is that the transformer- and CLIP-based backbones in our evaluation tend to be larger than the supervised convolutional baselines, so the observed advantages might simply reflect parameter count. Tables~\ref{tab:poporo} and \ref{tab:poporoin} include visual-encoder parameter counts to allow direct inspection, and Fig.~\ref{fig:scatter} plots ImageNet accuracy against POPORO and PoporoIN accuracy with circle area encoding model size. Parameter count is not unrelated to relatedness accuracy, but its association is weaker than that of architectural family or learning paradigm and is largely mediated by ImageNet accuracy, which itself scales approximately linearly with parameter count within a fixed architecture. At a fixed $\sim$86M ViT-B parameter budget, POPORO accuracy spans 50.50\% (DINOv1 ViT-B/16) to 79.00\% (CLIP LAION-2B ViT-B/16) --- a 28.5-percentage-point spread, indicating that factors beyond parameter count—such as pretraining paradigm, objective, and data—substantially affect relatedness accuracy; conversely, doubling the parameter count from CLIP LAION-2B ViT-L/14@336 (304M, POPORO 86.00\%) to ViT-H/14 (632M, POPORO 81.50\%) within the same paradigm leaves POPORO accuracy unchanged or slightly lower. Within-paradigm scaling effects are therefore smaller than cross-paradigm gaps at matched size, and the differences reported in Sections~V-B and V-C are not reducible to differences in parameter count.
 
\subsection{Taxonomic versus Thematic Relatedness}
The diagnostic metric $\Delta_{\text{CR-TR}}$ is positive for almost all models on POPORO. CLIP LAION-2B ViT-L/14@336 reaches CR=92.09\% and TR=78.92\% ($\Delta_{\text{CR-TR}}=+13.17$ percentage points); CLIP WIT-400M ViT-L/14 reaches CR=90.70\% and TR=77.84\% ($\Delta_{\text{CR-TR}}=+12.86$); and DINOv1 ViT-B/16 shows the largest gap on POPORO (CR=61.86\%, TR=37.30\%, $\Delta_{\text{CR-TR}}=+24.56$). On PoporoIN the trend is attenuated and several CLIP models reverse the order; CLIP WIT-400M ViT-L/14, in particular, has CR=74.00\% and TR=81.80\% ($\Delta_{\text{CR-TR}}=-7.80$). One possible explanation is that PoporoIN contains LLM-derived thematic pairs that may be more linguistically conventional than POPORO TR stimuli, making them easier for CLIP-style encoders pretrained on captioned web data. 
Taken together, the CR--TR gap indicates that current visual representations encode taxonomic relatedness more reliably than thematic or contextual relatedness on POPORO, paralleling the taxonomic--thematic asymmetry reported in human conceptual processing \cite{estes2011,mirman2017}; the PoporoIN reversal in selected CLIP models is consistent with the construction-bias caveat rather than with a representational reversal.
 
\subsection{Color- versus Shape-Matched Distractors}
The diagnostic metric $\Delta_{\text{CD-SD}}$ is positive for the majority of models on POPORO, indicating that shape-matched distractors mislead representations more often than color-matched ones. DINOv2 ViT-B/14 shows the largest gap on POPORO (CD=80.00\%, SD=61.74\%, $\Delta_{\text{CD-SD}}=+18.26$ percentage points), and DINOv1 ViT-B/8 shows a smaller but consistent gap (CD=62.35\%, SD=52.61\%, $\Delta_{\text{CD-SD}}=+9.74$). The pattern suggests that shape similarity is a stronger perceptual confound than color similarity under this triplet-ranking design, and may reflect that representations tuned to shape information can be misled when shape similarity is perceptual rather than semantic. This observation is related to but distinct from the shape-bias literature \cite{ritter2017cogpsych,geirhos2019,gavrikov2024}, which measures shape-versus-texture preference under cue conflict; a strong shape representation can simultaneously be human-aligned in cue-conflict classification and susceptible to shape-matched distractors in triplet ranking.
 
\subsection{Preliminary Observation on Generative Features}\label{sec:generative}
Guided Diffusion features at $t=90$, block 24, reach 49.00\% on POPORO and 60.50\% on PoporoIN; BigBiGAN reaches 47.25\% and 55.60\%. Despite low ImageNet accuracies (61.95\% and 60.80\%), both generative models are competitive with supervised ResNets on relatedness. Among the two evaluated generative representatives, Guided Diffusion exceeded BigBiGAN by +1.75 and +4.90 percentage points on POPORO and PoporoIN, respectively. Diffusion accuracy varies with timestep (POPORO accuracy 46.00\%--50.50\% across the sweep, with intermediate timesteps around $t=100$--$200$ showing relatively stronger performance). 
 
\subsection{Qualitative Analysis}
Fig.~\ref{fig:qualitative} grounds the quantitative patterns from Sections~V-F and V-G in concrete examples from the strongest model on each benchmark. The agreement cells (top row) show that the model correctly places taxonomically and thematically related targets closer to the query than the distractor --- CR examples include POPORO category pairs and PoporoIN class pairs such as \emph{microphone--loudspeaker}, \emph{harmonica--guitar}, and \emph{rifle--revolver}, while TR examples include POPORO co-occurrence pairs like \emph{bathtub--faucet} and PoporoIN complementary pairs such as \emph{corkscrew--wine bottle} and \emph{drum--drumsticks}. The disagreement cells (bottom row) instantiate the CD-SD asymmetry reported numerically: in CD trials, a salient color match (e.g., red helmet versus red flower) can override the semantically correct target, but the failure is more pronounced in SD trials, where shape-matched distractors (e.g., \emph{hamster--squirrel monkey}; \emph{horseshoe--horse}) override semantically correct related targets that the model otherwise embeds nearby. The arrangement therefore confirms visually that the strongest model still inherits the CR-over-TR and SD-over-CD weaknesses summarized by the diagnostic gap metrics, rather than producing a qualitatively different error profile.

%% file: sec_6_discussion.tex
\section{Discussion}\label{sec:discussion}
 
\subsection{Principal Findings}
Three findings stand out under the evaluated benchmarks. First, transformer-based and vision--language pretrained representations align more closely with object-level relatedness than convolutional and vision-only representations, and the gap is not fully captured by ImageNet accuracy. Relatedness accuracy varies systematically with architectural family and learning paradigm, while parameter count and ImageNet accuracy within a fixed family are weaker predictors. Second, both axes of the annotation scheme exhibit consistent asymmetries: the CR--TR gap mirrors the taxonomic--thematic asymmetry documented in human conceptual processing, and the CD--SD gap exposes shape similarity as the more potent perceptual confound. Third, classification accuracy and relatedness accuracy are not reducible to one another, so benchmarks targeting recognition do not characterize this representational property; among the two evaluated generative representatives, both perform comparably to supervised CNNs despite their lower ImageNet accuracy, which is consistent with this dissociation.
 
\subsection{Why the Protocol Is Diagnostic}
The triplet-ranking protocol is deliberately minimal. By using cosine similarity on frozen pretrained features and avoiding any fine-tuning, the protocol measures the representation as it exists at the end of pretraining, free of confounds introduced by adaptation choices. The two-axis annotation scheme separates effects that would otherwise be entangled: a representation can be strong on taxonomic relations and weak on thematic ones (DINOv2 on POPORO), or be misled chiefly by shape rather than color (the same model on POPORO), and these patterns are visible because the four cells are scored independently. The diagnostic gap metrics package these patterns into a compact summary that does not require additional experiments.
 
\subsection{Architectural and Paradigm Interpretations}
The architectural and paradigm patterns admit consistent, if not yet causal, interpretations. One possible interpretation of the transformer advantage is that transformer architectures provide global interactions that may preserve object-level co-occurrence cues more readily than local convolutional architectures; the supervised RN-50/RN-101 results, where increased convolutional capacity does not lift POPORO accuracy out of the near-chance range, are consistent with this pattern. We do not, however, analyze attention maps or layer dynamics in this work, so this remains an interpretation rather than a causal conclusion. Similarly, CLIP-style image--text pretraining may encourage visual features to align with object co-occurrence and concept associations expressed in language, but this remains confounded with data scale and data diversity in the present comparison. Both interpretations are consistent with our observations but neither is causally established by the present data; pretraining objective, data scale, and architecture remain entangled.
 
\subsection{Relation to Prior Work}
Our benchmarks complement, rather than replace, existing relation benchmarks. VRD, Visual Genome, and scene graph generation \cite{lu2016vrd,krishna2017visualgenome,xu2017scenegraph} measure within-image predicate structure; spatial reasoning benchmarks \cite{kamath2023whatsup,liu2023vsr} measure spatial predicates; and VLM compositionality benchmarks \cite{yuksekgonul2023aro,hsieh2023sugarcrepe,thrush2022winoground,tong2024mmvp,hua2024mmcomposition} probe through image--text matching. Our protocol fills the niche where the question is whether two object concepts are encoded as related in a purely visual representation, without language as an intermediary and without spatial structure within a scene. The Concept Association Bias study \cite{yamada2022cab} and our protocol are complementary perspectives on the same VLMs: the former diagnoses a bag-of-concepts failure mode, the latter quantifies how reliably the representation places related objects together.
 
\subsection{Practical Implications}
For model selection in retrieval, scene-level reasoning, or analogical search tasks where object-level relatedness matters, ImageNet accuracy alone may be insufficient as a quality indicator. Within a fixed paradigm and architecture, ImageNet accuracy correlates with relatedness accuracy, but across paradigms the relationship breaks: a CLIP backbone with lower ImageNet accuracy can encode relatedness more reliably than a supervised backbone with higher accuracy. Model selection for tasks that depend on object-level relatedness should therefore weight architectural family and learning paradigm rather than relying on ImageNet accuracy alone.
 
\subsection{Limitations and Future Work}\label{sec:limitations-future}
Several limitations qualify the present results. Cross-paradigm comparisons combine learning paradigm with data scale and training-set composition (ImageNet-1k for supervised and most SSL models, LVD-142M for DINOv2, WIT-400M for OpenAI CLIP, LAION-2B for OpenCLIP), so the same-architecture, near-equal-ImageNet-accuracy CLIP-RN50 versus VICReg-RN50 comparison reported in Section~V-C contextualizes but does not remove this confound. PoporoIN is built from ImageNet validation images and its candidate pairs were proposed by LLMs and filtered by author review without external inter-annotator agreement, so part of the CLIP advantage observed on PoporoIN may reflect language-aligned pair construction or web-corpus overlap with ImageNet rather than purely visual representation quality; the matched POPORO comparison, which predates the LLM era and uses stylized cutouts unlikely to appear verbatim in web crawls, mitigates but does not eliminate this concern. The protocol applies a single cosine readout on a penultimate or final feature, uses the visual encoder of CLIP only, and reports descriptive single-point accuracies without bootstrap intervals or paired tests; alternative readouts, zero-shot text-based ranking, and uncertainty quantification could change details of the ranking. Modern multimodal foundation models (e.g., GPT-4V, Gemini, LLaVA-NeXT, SigLIP, EVA-CLIP) are also outside the present evaluation, and our benchmarks treat object-level relatedness as a static pairwise property rather than a context-dependent judgment.
 
Future work follows directly from these limitations along three axes. First, \emph{benchmark validation}, including external human inter-annotator agreement on PoporoIN and protocol-aligned human triplet-ranking accuracy on POPORO that closes the gap with the original psychometric ratings, would calibrate the absolute scale on which models are compared. Second, \emph{protocol broadening}, including layer and metric ablations, learned probes, and zero-shot text-encoder baselines under matched conditions, would quantify how much of the reported structure is specific to the cosine-on-final-feature readout. Third, the consistent direction of $\Delta_{\text{CR-TR}}>0$ for most models on POPORO suggests \emph{pretraining objectives targeted at thematic relatedness} as a candidate intervention: future strategies that explicitly model object co-occurrence, temporal continuity, or grounded interaction context could in principle reduce the CR--TR gap, and the diagnostic gap metrics introduced here are set up so that such hypotheses can be tested without conflating relatedness gains with classification gains.

%% file: sec_7_conclusion.tex
\section{Conclusion}\label{sec:conclusion}

We introduced an image-only triplet-ranking protocol for evaluating object-level semantic relatedness in pretrained vision representations. We instantiated the protocol on two test beds: POPORO \cite{kovalenko2012poporo}, an existing psychologically validated stimulus set repurposed for representation evaluation, and PoporoIN, a newly constructed and curated 1{,}000-triplet ImageNet-based benchmark extension. By organizing each triplet along two orthogonal axes that distinguish taxonomic from thematic targets and color- from shape-matched distractors, the protocol exposes representational structure that recognition accuracy alone does not fully predict. Across twenty pretrained models, transformer-based and vision--language pretrained representations align more closely with the benchmark-defined relatedness labels (and, on POPORO specifically, with human-defined relatedness) than convolutional and vision-only counterparts; taxonomic relatedness is encoded more reliably than thematic relatedness; and shape similarity is the more potent perceptual confound. The protocol, the PoporoIN benchmark extension, and the diagnostic gap metrics together provide a compact diagnostic instrument with which the cognitive-psychology distinction between taxonomic and thematic relations can be carried into the evaluation of visual representations.